\documentclass[letterpaper, 10 pt, conference]{ieeeconf}  
\usepackage{array}
\usepackage{times}  
\usepackage{helvet}  
\usepackage{courier}  
\usepackage[hyphens]{url}  
\usepackage{graphicx} 
\usepackage{multirow}
\usepackage{booktabs}
\usepackage[table]{xcolor} 
\usepackage{xcolor}
\usepackage{caption} 
\usepackage{amsthm}
\usepackage{algorithm}
\usepackage{algorithmic}
\usepackage{adjustbox}
\usepackage{newfloat}
\usepackage{listings}
\usepackage{amsfonts}
\usepackage{makecell}
\usepackage{float}
\usepackage{amsmath}
\usepackage{amssymb}
\usepackage{mathtools}
\usepackage{bm}
\usepackage[capitalize,noabbrev]{cleveref}

\theoremstyle{plain}

\theoremstyle{definition}

\theoremstyle{remark}

\newcommand{\method}{StaKe}

\IEEEoverridecommandlockouts                              

\title{\LARGE \bf
Improving Vision-Language-Action Model Fine-Tuning with Structured Stage and Keyframe Supervision
}

\author{Yuan Xu$^{1,2}$,\quad Yixiang Chen$^{1,2}$,\quad Kai Wang$^{2}$,\quad Jiabing Yang$^{1,2}$ \\[3pt]
Peiyan Li$^{1,2}$,\quad Qisen Ma$^{1,2}$,\quad Yan Huang$^{1,2,3\dag}$,\quad Liang Wang$^{1,2\dag}$
\thanks{$\dag$ Corresponding author}
\thanks{$^{1}$School of Artificial Intelligence, University of Chinese Academy of Sciences. $^{2}$New Laboratory of Pattern Recognition (NLPR), State Key Laboratory of Multimodal Artificial Intelligence Systems (MAIS), Institute of Automation, Chinese Academy of Sciences. $^{3}$FiveAges.}
\thanks{Emails: {\tt \{yuan.xu, yhuang, wangliang\}@nlpr.ia.ac.cn}}
}
\begin{document}


\maketitle
\thispagestyle{empty}
\pagestyle{empty}

\begin{abstract}
Vision-Language-Action (VLA) models have shown strong potential for generalizable robotic manipulation. During fine-tuning, however, action supervision applies equally across all timesteps, without structured supervision on which manipulation stage the robot is in or what the next gripper-event target should be. This causes failures to concentrate around challenging gripper-event transitions.
To address this, we propose \textbf{\method{}}, a plug-in auxiliary supervision framework that automatically derives two complementary signals from demonstration gripper states without manual annotation: a stage classifier that identifies the current manipulation stage, and a keyframe predictor that estimates the target joint action at the next gripper transition. Both are modeled as lightweight auxiliary heads that enrich the learned representations during training, while leaving the base VLA policy architecture and inference loop unchanged.
Experiments on bimanual simulation and single-arm Franka real-robot tasks show that \method{} consistently improves success rates (relative gains of 14\% and 56\%, respectively), with larger improvements on longer-horizon tasks that involve more gripper-event transitions. Ablation studies validate each design choice, and qualitative analysis confirms that the learned representations faithfully track manipulation stages.
These results indicate that structured supervision is an effective and general strategy for enhancing VLA fine-tuning in long-horizon manipulation.
Project website: https://hi-yuanxu.github.io/StaKe-Web/.
\end{abstract}




\section{INTRODUCTION}

Enabling robots to interpret open-ended language instructions and execute diverse manipulation skills is a long-standing aspiration of embodied intelligence~\cite{zhu2024vision}. The rapid progress of Vision-Language Models (VLMs)~\cite{awadalla2023openflamingo,liu2024visual}, which acquire rich cross-modal representations from internet-scale data, has opened a promising path toward this goal. By grounding VLM representations in physical action spaces, Vision-Language-Action (VLA) models~\cite{brohan2022rt1,brohan2023rt2,driess2023palme,kim2024openvla,octo2024} cast robotic manipulation as a language-conditioned generation problem and learn end-to-end mappings from visual observations and instructions to executable actions. Pre-trained on large-scale robotic datasets~\cite{o2024open,fang2024rh20t} and fine-tuned on downstream tasks, recent VLA policies such as $\pi_0$~\cite{black2024pi0} and $\pi_{0.5}$~\cite{physicalintelligence2025pi05} have shown remarkable generalization across a wide range of manipulation scenarios.

\begin{figure}[t]
    \centering
    \includegraphics[width=0.95\linewidth]{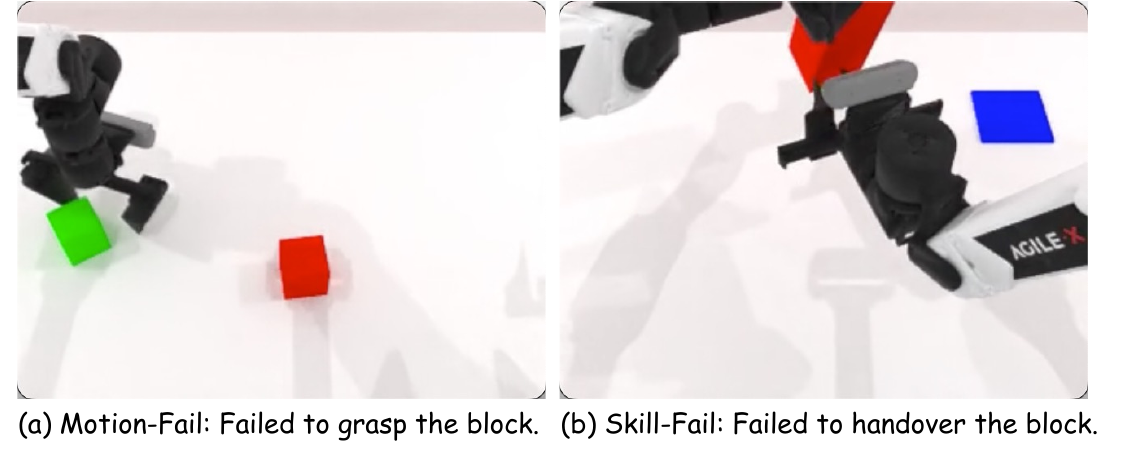}
    \caption{\textbf{Manipulation failures concentrate around gripper-state transitions.} A \emph{motion-fail} occurs when the robot fails to reach and grasp the target during free-space motion, while a \emph{skill-fail} occurs when the robot grasps successfully but fails during post-grasp skill execution (e.g., handover, stacking, rotation).}
    \label{fig:intro_failure_decomposition}
\end{figure}

When fine-tuning these models for real-world deployment, current practice relies on a continuous action loss applied uniformly over the entire trajectory~\cite{openvla_oft2025,bharadhwaj2023roboagent}. Yet a manipulation trajectory is not homogeneous, but typically consists of several structured stages, such as free-space motion toward an object and contact-constrained skill execution after grasping, with gripper-state transitions serving as critical stage boundaries. Despite this structure, the uniform action loss treats all timesteps equally and provides no structured supervision for either the current manipulation stage or the upcoming transition target, leaving the policy to discover this structure entirely on its own. As illustrated in Fig.~\ref{fig:intro_failure_decomposition}, failures tend to concentrate around these stage boundaries, manifesting as either a \emph{motion-fail} when the robot fails to reach and grasp the target during free-space motion, or a \emph{skill-fail} when it grasps successfully but errs during post-grasp execution such as handover, stacking, or rotation.

\begin{figure}[t]
  \centering
  \includegraphics[width=0.98\linewidth]{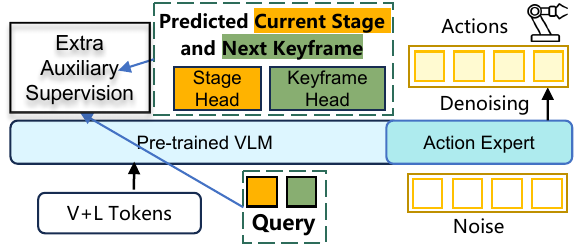}
  \caption{\textbf{Auxiliary supervision introduced by \method{}.} Two learnable query tokens and their corresponding heads (highlighted) are added to a VLA model, providing stage classification and keyframe prediction signals during training without altering the action expert or inference loop.}
  \label{fig:method_comparison}
\end{figure}

To address this, we propose \textbf{\method{}} (\textbf{Sta}ge and \textbf{Ke}yframe supervision), a structured supervision framework for VLA fine-tuning (Fig.~\ref{fig:method_comparison}). \method{} introduces two auxiliary training signals: \emph{Stage Supervision~(SS)}, which classifies each timestep into motion or skill stages, and \emph{Keyframe Supervision~(KS)}, which predicts the joint action at the next gripper-transition keyframe. Both labels are derived automatically from demonstration gripper states without manual annotation, and the auxiliary heads are active only during training, leaving the inference loop entirely unchanged.

We validate \method{} through extensive experiments on RoboTwin2.0~\cite{chen2025robotwin} dual-arm simulation tasks and real-world Franka single-arm tasks. Compared with the base VLA models, \method{} consistently improves task success rates across both short-horizon and long-horizon manipulation scenarios. Ablation studies confirm the importance of each component, and detailed analysis of per-keyframe success rates and episode-level execution traces provides insight into the effectiveness and locality of the auxiliary supervision.

Our primary contributions include:
\begin{itemize}
\item We propose \method{}, a plug-in structured supervision framework for VLA fine-tuning that introduces two complementary signals---Stage Supervision~(SS) and Keyframe Supervision~(KS)---to improve fine-tuning performance while leaving the inference loop unchanged.
\item We design a fully automatic pipeline that extracts stage labels and keyframe targets from demonstration gripper states, enabling structured supervision without any manual annotation.
\item We validate \method{} on bimanual simulation and single-arm real-robot tasks, demonstrating consistent success-rate improvements across diverse scenarios, with larger gains on longer-horizon tasks that involve more gripper-state transitions.
\end{itemize}

\section{RELATED WORKS}

\subsection{Vision-Language-Action Models}
Vision-Language-Action (VLA) models map visual observations and language instructions directly to executable robot actions by coupling pre-trained vision-language backbones with action prediction modules. Early architectures such as RT-1~\cite{brohan2022rt1}, RT-2~\cite{brohan2023rt2}, and PaLM-E~\cite{driess2023palme} established the paradigm of scaling Transformer-based models on large robotic datasets for multi-task generalization. A subsequent wave of open-source efforts~\cite{kim2024openvla,octo2024,li2024roboflamingo,li2024cogact} broadened accessibility and demonstrated that adapting vision-language foundation models yields robust manipulation capabilities across diverse settings. More recently, the field has advanced along several axes: flow-matching and diffusion-based action generation~\cite{black2024pi0,physicalintelligence2025pi05,liu2024rdt}, heterogeneous cross-embodiment pre-training~\cite{cheang2024gr2,wang2024hpt}, and spatial-aware or hybrid architectures~\cite{qu2025spatialvla,liu2025hybridvla}. Despite this progress, existing VLA fine-tuning pipelines rely on a uniform continuous action loss and do not explicitly encode the stage structure of manipulation trajectories. Our \method{} addresses this gap by augmenting the training loss with lightweight stage and keyframe supervision, without modifying the policy architecture or inference procedure.

\subsection{Fine-Tuning Strategies for VLA Models}
Deploying VLA models on downstream tasks requires fine-tuning recipes balancing performance, efficiency, and generality. One line revisited action representations and decoding within supervised fine-tuning: OpenVLA-OFT~\cite{openvla_oft2025} optimized parallel decoding and continuous action regression, while FAST~\cite{pertsch2025fast} introduced frequency-space action tokenization for autoregressive policies. Concurrently, TinyVLA~\cite{wen2024tinyvla} and DeeR-VLA~\cite{yue2024deervla} reduced compute via lightweight backbones or dynamic inference depth. A separate line integrated reinforcement learning into fine-tuning: ConRFT~\cite{chen2025conrft} augmented behavior cloning with consistency-based RL, and iRe-VLA~\cite{guo2025irevla} adopted iterative RL updates for online adaptation. These methods primarily modify architecture, action parameterization, or the optimization objective. In contrast, \method{} adds stage- and keyframe-level auxiliary losses to a standard fine-tuning loop without modifying the policy architecture or inference procedure, and is therefore complementary to the improvements above.

\subsection{Structured Supervision for Manipulation}
A growing body of work improves manipulation policies by learning attention toward temporally critical states rather than treating all timesteps uniformly. Along the temporal axis, one line predicts sparse keyframes or waypoints capturing a trajectory's decisive moments: PerAct~\cite{shridhar2023perceiver} and RVT~\cite{goyal2023rvt} predict 6-DoF keyposes from 3D observations; Act3D~\cite{gervet2023act3d} and Coarse-to-Fine 3D Keyframe Transporter~\cite{chen2025c2fkeyframe} refine keyframe localization via hierarchical sampling; AWE~\cite{shi2023awe} and PIVOT-R~\cite{zhang2024pivotr} extract or rank waypoints to shorten the decision horizon. Complementary to temporal selection, a spatial line operates in the keypoint domain, where KAT~\cite{dipalo2024kat} represents manipulation via keypoint action tokens and ReKep~\cite{huang2024rekep} formulates relational keypoint constraints optimizable at inference. A third line exploits skill-level structure via task segmentation, semantic skill libraries, or auxiliary contrastive objectives~\cite{kou2024kisa,prime2024,atomskill2025,dexskills2024,ma2024sigmaagent}. While these approaches demonstrate the value of structured supervision, they generally rely on manually defined keyframes, task-specific annotations, or dedicated architectures. \method{} instead automatically derives all labels from demonstration gripper states and operates as lightweight auxiliary losses, preserving the pre-trained VLA backbone and inference procedure unchanged.

\section{METHOD}
\label{sec:method}

\begin{figure*}[t]
  \vspace{4pt}
  \centering
  \includegraphics[width=0.7\textwidth]{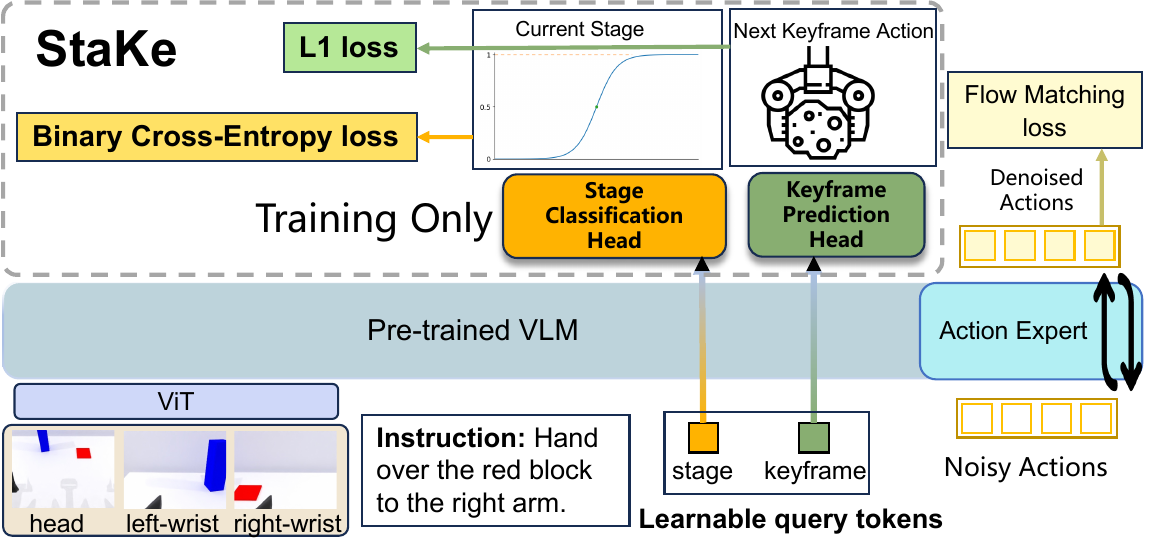}
  \caption{\textbf{Overview of the \method{} framework.} Two learnable query tokens are appended to the pre-trained VLM backbone, each feeding a lightweight auxiliary head: a stage classification head trained with binary cross-entropy loss and a keyframe prediction head trained with L1 loss. Both heads are active only during training (marked ``Training Only'') and jointly optimized with the flow-matching policy loss. At inference time, the query tokens remain in the sequence but the auxiliary heads are not invoked.}
  \label{fig:model_overview}
\end{figure*}

In this section, we introduce the details of \method{}. We first present the base VLA architecture in Sec.~\ref{subsec:prelim}. Subsequently, we describe the automatic extraction of stage labels and keyframe targets from demonstrations in Sec.~\ref{subsec:stage_labeling} and Sec.~\ref{subsec:keyframe_extraction}. Finally, we present the model architecture and implementation details in Sec.~\ref{subsec:arch_impl}. Fig.~\ref{fig:model_overview} provides an overview of the complete framework.

\subsection{Preliminaries: $\pi_{0.5}$ VLA Architecture}
\label{subsec:prelim}

We build upon the $\pi_{0.5}$ VLA architecture~\cite{physicalintelligence2025pi05}, which consists of a vision encoder $f_{\text{enc}}$, a pre-trained language model backbone, and a flow-matching action expert. At each timestep $t$, the model receives a visual observation $I_t$ and a language instruction $l$, and produces an action chunk $\hat{a}_{t:t+H-1}$ of horizon $H$ in joint space. The vision encoder extracts visual tokens $h_{\text{img}} = f_{\text{enc}}(I_t)$, and the language instruction is tokenized into $h_l$. The language model backbone processes the concatenation of $h_{\text{img}}$ and $h_l$ to produce a latent representation $z_t$, from which the flow-matching action expert generates the action chunk.

\noindent
\textbf{Training.}
The action expert is trained via flow matching~\cite{chi2023diffusion}. Let $a_0$ denote the ground-truth action chunk from a demonstration and $\epsilon \sim \mathcal{N}(0, I)$ a noise sample. The forward process constructs a noised action $a_\tau$ at noise level $\tau \in [0,1]$:
\begin{equation}
a_\tau = (1 - \tau)\, a_0 + \tau\, \epsilon.
\label{eq:forward}
\end{equation}
The target velocity field is defined as $u = a_0 - \epsilon$. A policy network $\hat{v}_\theta$ is trained to predict this velocity, yielding the flow-matching loss:
\begin{equation}
L_{\text{policy}} = \mathbb{E}_{\tau, \epsilon} \left\| \hat{v}_\theta(a_\tau, \tau, z_t) - u \right\|^2.
\label{eq:flow_loss}
\end{equation}

\noindent
\textbf{Inference.}
At test time, the action expert generates an action chunk by solving the reverse ODE from $a_1 \sim \mathcal{N}(0, I)$:
\begin{equation}
a_{\tau - \Delta\tau} = a_{\tau} + \Delta\tau \cdot \hat{v}_\theta(a_{\tau}, \tau, z_t),
\label{eq:denoise}
\end{equation}
iterating from $\tau=1$ to $0$ with step size $\Delta\tau$. The final $a_0$ is the predicted action chunk executed on the robot.

\subsection{Stage Label Extraction}
\label{subsec:stage_labeling}

To provide the policy with structured supervision beyond the continuous action loss, we automatically extract two complementary signals from demonstration gripper states: stage labels that capture the stage-level structure of a trajectory, and keyframe targets that anchor the policy to critical gripper-transition frames. Both signals require no manual annotation. We describe the stage labeling procedure here and the keyframe extraction in Sec.~\ref{subsec:keyframe_extraction}.

A manipulation trajectory naturally alternates between free-space motion and contact-constrained skill execution, with gripper events marking the boundaries. We formalize this by parsing the gripper open/close sequence of each demonstration. Let $M$ denote the number of arms ($M{=}1$ for single-arm, $M{=}2$ for bimanual). A frame is considered closed whenever \emph{any} of the $M$ grippers is closed, since a closed gripper indicates that the robot is actively executing a contact skill. Let $[t_1, t_2]$ denote each such interval. To capture the pre-contact preparation and post-contact settling phases, we expand each interval by a margin of $K$ frames:
\begin{equation}
a = \max(1,\, t_1 - K), \quad b = \min(T,\, t_2 + K).
\label{eq:stage_expand}
\end{equation}
All frames $t \in [a, b]$ are labeled as \emph{skill} ($s_t = 1$), and the remaining frames as \emph{motion} ($s_t = 0$). Algorithm~\ref{alg:stage_label} summarizes this procedure. The expansion margin $K$ is a tunable hyperparameter whose sensitivity we study in the ablation experiments.

\begin{algorithm}[ht]
\caption{Stage Label Extraction}\label{alg:stage_label}
\begin{algorithmic}[1]
\REQUIRE Gripper states $g^{(m)}_t$ for $M \in \{1,2\}$ arms ($1 \le t \le T$, $1 \le m \le M$), expansion margin $K$
\STATE Initialize $s_t = 0$ (motion) for $t=1,\dots,T$.
\STATE Identify all intervals $[t_1,t_2]$ where $\exists\, m: g^{(m)}_t = \text{closed}$.
\FOR{each closed interval $[t_1,t_2]$}
  \STATE $a=\max(1, t_1 - K)$, $b=\min(T, t_2 + K)$.
  \FOR{$t = a$ to $b$}
    \STATE $s_t \gets 1$ (skill).
  \ENDFOR
\ENDFOR
\RETURN Stage labels $s_{1:T}$.
\end{algorithmic}
\end{algorithm}

\subsection{Keyframe Target Extraction}
\label{subsec:keyframe_extraction}

Complementary to the stage labels that provide stage-level supervision, keyframe targets offer frame-level goal signals anchored to gripper-transition events. The extraction proceeds in two steps. First, we scan the trajectory for frames where any of the $M$ grippers changes state (open $\rightarrow$ closed or closed $\rightarrow$ open). When $M{=}2$, transitions from both arms are detected independently and merged into a single ordered set, since the action vector covers both arms jointly (14 dimensions for two 7-DoF arms). We record the full joint configuration at each transition frame as a keyframe. Second, we assign a per-frame supervision target: for each timestep $t$, we find the nearest upcoming transition frame $t_{\text{next}}$ from the merged set and set $q^{\text{kf}}_t = q_{t_{\text{next}}}$. We additionally include the terminal frame $(T, q_T)$ as the final keyframe, so that timesteps after the last gripper transition point toward the task-completion configuration. In this way, every frame in the trajectory receives a keyframe label pointing toward its next critical gripper event, regardless of which arm triggers it. Algorithm~\ref{alg:keyframe_extraction} summarizes the complete procedure.

\begin{algorithm}[ht]
\caption{Keyframe Target Extraction}\label{alg:keyframe_extraction}
\begin{algorithmic}[1]
\REQUIRE Gripper states $g^{(m)}_t$ for $M \in \{1,2\}$ arms, joint states $q_t$, $t=1\dots T$, $1 \le m \le M$
\STATE \textbf{Step 1: Identify transition frames (per-arm, then merge)}
\STATE $E \gets []$
\FOR{$m=1$ to $M$}
  \FOR{$t=1$ to $T-1$}
    \IF{$g^{(m)}_t \neq g^{(m)}_{t+1}$}
      \STATE Append $(t+1, q_{t+1})$ to $E$.
    \ENDIF
  \ENDFOR
\ENDFOR
\STATE Sort $E$ by frame index; remove duplicates.
\STATE Append $(T, q_T)$ to $E$. \hfill $\triangleright$ terminal frame as final keyframe
\STATE \textbf{Step 2: Assign per-frame keyframe labels}
\STATE $j \gets 1$
\FOR{$t=1$ to $T$}
  \WHILE{$j < |E|$ \AND $E[j].\text{frame} < t$}
    \STATE $j \gets j + 1$
  \ENDWHILE
  \STATE $q^{\text{kf}}_t \gets E[\min(j, |E|)].\text{joint}$
\ENDFOR
\RETURN Per-frame keyframe targets $q^{\text{kf}}_{1:T}$.
\end{algorithmic}
\end{algorithm}

The next gripper-transition keyframe serves as a semantically meaningful goal for each timestep, since the robot must achieve precise alignment at the moment of grasping or releasing. This event-driven definition is more task-relevant than a fixed-horizon target and generalizes across different manipulation scenarios.

\subsection{Model Architecture and Implementation Details}
\label{subsec:arch_impl}

\noindent
\textbf{Learnable Query Tokens.}
To extract auxiliary predictions without modifying the base architecture, we introduce two learnable query tokens, $e^{\text{ss}}$ and $e^{\text{kf}}$, which are appended to the input sequence of the language model backbone alongside $h_{\text{img}}$ and $h_l$ (see Fig.~\ref{fig:model_overview}). The backbone processes all tokens jointly, producing contextualized representations for both the original tokens and the two queries. Let $\tilde{z}^{\text{ss}}_t$ and $\tilde{z}^{\text{kf}}_t$ denote the output representations corresponding to the stage and keyframe query tokens at timestep $t$, respectively.

\noindent
\textbf{Auxiliary Heads.}
The stage head is a small MLP that maps $\tilde{z}^{\text{ss}}_t$ to a scalar prediction $\hat{s}_t = \text{MLP}_{\text{ss}}(\tilde{z}^{\text{ss}}_t) \in [0, 1]$, representing the probability of the current timestep being in the skill stage. It is trained with binary cross-entropy:
\begin{equation}
L_{\text{stage}} = -\frac{1}{T}\sum_{t=1}^{T} \left[ s_t \log \hat{s}_t + (1 - s_t) \log (1 - \hat{s}_t) \right].
\label{eq:stage_loss}
\end{equation}
The keyframe head is another MLP that maps $\tilde{z}^{\text{kf}}_t$ to a joint-space prediction $\hat{q}^{\text{kf}}_t = \text{MLP}_{\text{kf}}(\tilde{z}^{\text{kf}}_t)$, trained with L1 loss against the extracted keyframe target:
\begin{equation}
L_{\text{kf}} = \frac{1}{T}\sum_{t=1}^{T} \left\| \hat{q}^{\text{kf}}_t - q^{\text{kf}}_t \right\|_1.
\label{eq:kp_loss}
\end{equation}

\noindent
\textbf{Training.}
The total training loss combines the flow-matching policy loss with the two auxiliary losses (Fig.~\ref{fig:model_overview}):
\begin{equation}
L = L_{\text{policy}} + \lambda_s L_{\text{stage}} + \lambda_k L_{\text{kf}},
\label{eq:total_loss}
\end{equation}
where $\lambda_s$ and $\lambda_k$ are empirically set to $0.01$ and $0.01$ to balance the scale difference between the auxiliary losses and the flow-matching policy loss, and kept fixed across all tasks. Both auxiliary heads add negligible computation overhead. We normalize keyframe target actions per-dimension to stabilize training.

\noindent
\textbf{Inference.}
At test time, the auxiliary heads are not invoked (marked ``Training Only'' in Fig.~\ref{fig:model_overview}), and the inference procedure remains identical to the base $\pi_{0.5}$ policy: given observation $I_t$ and instruction $l$, the action expert generates an action chunk $\hat{a}_{t:t+H-1}$ via iterative denoising as in Eq.~\eqref{eq:denoise}.

\section{EXPERIMENTS}
We conducted comprehensive experiments to answer the following questions: \textbf{Q1.} Does \method{} improve manipulation success rates compared with the base $\pi_{0.5}$ policy and other strong baselines in bimanual simulation?
\textbf{Q2.} How does each component of \method{} contribute, and how sensitive is performance to key design choices?
\textbf{Q3.} Do the learned stage and keyframe predictions align with the intended trajectory structure?
\textbf{Q4.} Do the gains transfer to real-robot tasks, and how does performance compare at each keyframe stage throughout the manipulation process?

\begin{table*}[t]
\vspace{4pt}
\centering
\caption{\textbf{Success rates (\%) on 10 RoboTwin~2.0 bimanual tasks (\textbf{Q1}).} Tasks are grouped by the number of keyframes (\#KF). DP, ACT, and RDT results are from the official benchmark~\cite{chen2025robotwin}. Best results per task are in \textbf{bold}.}
\label{tab:sim_results}
\begin{tabular}{l c ccc cc}
\toprule
Task & \#KF & DP~\cite{chi2023diffusion} & ACT~\cite{zhao2023act} & RDT~\cite{liu2024rdt} & $\pi_{0.5}$~\cite{physicalintelligence2025pi05} & \method{} \\
\midrule
Adjust Bottle       & 2 & \textbf{97} & \textbf{97} & 81 & 90 & 96 \\
Click Alarmclock    & 2 & 61 & 32 & 61 & 63 & \textbf{76} \\
Lift Pot            & 2 & 39 & \textbf{88} & 72 & 22 & 36 \\
Move Can Pot        & 2 & 39 & 22 & 25 & 42 & \textbf{46} \\
Open Laptop         & 2 & 49 & 56 & 59 & 78 & \textbf{80} \\
Rotate Qrcode       & 2 & 13 &  1 & \textbf{50} & 44 & \textbf{50} \\
\midrule
Put Object Cabinet  & 3 & \textbf{42} & 15 & 33 & 24 & 29 \\
\midrule
Handover Block      & 4 & 10 & 42 & 45 & 55 & \textbf{63} \\
Place Burger Fries  & 4 & 72 & 49 & 50 & 64 & \textbf{76} \\
\midrule
Stack Blocks Two    & 5 &  7 & 25 & 21 & 42 & \textbf{46} \\
\midrule
\rowcolor{gray!10}
Average             & -- & 42.9 & 42.7 & 49.7 & 52.4 & \textbf{59.8} \\
\bottomrule
\end{tabular}
\end{table*}

\subsection{Experimental Setup}
\label{subsec:setup}

\noindent
\textbf{Simulation environment.}
We evaluate on the RoboTwin~2.0 benchmark~\cite{chen2025robotwin}, a bimanual manipulation suite built on the SAPIEN simulator. We select 10 tasks spanning four levels of keyframe complexity: six 2-keyframe tasks (\emph{Adjust Bottle}, \emph{Click Alarmclock}, \emph{Lift Pot}, \emph{Move Can Pot}, \emph{Open Laptop}, \emph{Rotate Qrcode}), one 3-keyframe task (\emph{Put Object Cabinet}), two 4-keyframe tasks (\emph{Handover Block}, \emph{Place Burger Fries}), and one 5-keyframe task (\emph{Stack Blocks Two}). At each evaluation episode, a language instruction is sampled randomly from the task-specific instruction pool provided by the simulator. We use the \texttt{demo\_clean} setting with 50 demonstrations per task.

\noindent
\textbf{Training details.}
All models undergo full-model post-training (updating both the VLM backbone and the action expert) on 4$\times$NVIDIA H20 GPUs for 20\,000 steps with a batch size of 64 and a learning rate of $5\times10^{-5}$. The action chunk horizon is $H{=}50$. For \method{}, the stage-label expansion margin is set to $K{=}25$ (about 1\,s, covering the pre-grasp and object-alignment phase) and the auxiliary loss weights follow Eq.~\eqref{eq:total_loss}.

\noindent
\textbf{Evaluation protocol.}
Each task is evaluated over 100 episodes. We report the success rate determined by the benchmark's built-in checker.

\noindent
\textbf{Baselines.}
Our primary baseline is $\pi_{0.5}$~\cite{physicalintelligence2025pi05}, a state-of-the-art VLA policy, fine-tuned with the same data and training recipe but without auxiliary losses. We additionally compare against three representative manipulation methods: Diffusion Policy (DP)~\cite{chi2023diffusion}, ACT~\cite{zhao2023act}, and RDT-1B~\cite{liu2024rdt}, using results reported by the official RoboTwin~2.0 benchmark~\cite{chen2025robotwin}.

\subsection{Simulation Results (\textbf{Q1})}

Table~\ref{tab:sim_results} reports success rates across all 10 tasks. We highlight three findings.

\noindent
\textbf{Overall performance.}
\method{} achieves the highest average success rate (59.8\%) among all five methods, outperforming DP (42.9\%), ACT (42.7\%), RDT (49.7\%), and $\pi_{0.5}$ (52.4\%). It obtains the best or tied-best result on 7 of the 10 tasks. On the three tasks where other methods lead, \method{} still improves over $\pi_{0.5}$. These results confirm that the lightweight auxiliary losses strengthen the base VLA policy without sacrificing generality.

\noindent
\textbf{Effect of auxiliary supervision.}
Compared with $\pi_{0.5}$, which shares the same architecture and training recipe, \method{} improves success on all 10 tasks ($+7.4$ pp on average). This gain arises solely from the auxiliary stage and keyframe losses, with no modification to model architecture or inference cost. The improvement holds both where $\pi_{0.5}$ already performs well (e.g., \emph{Adjust Bottle} $90{\to}96$) and where it struggles (e.g., \emph{Lift Pot} $22{\to}36$), confirming that the structured supervision effectively complements the base policy loss regardless of task difficulty.

\noindent
\textbf{Notable gains on multi-keyframe tasks.}
On the two 4-keyframe tasks, \method{} improves $\pi_{0.5}$ by $+8$ and $+12$ points respectively. On the most complex 5-keyframe task (\emph{Stack Blocks Two}), \method{} reaches 46\% vs.\ 42\% for $\pi_{0.5}$ and substantially outperforms DP, ACT, and RDT. As the number of keyframes grows, each gripper-transition stage becomes a potential failure point; \method{} directly addresses this by anchoring the policy at every such stage.

\subsection{Ablation Studies (\textbf{Q2})}

We conduct ablation experiments on a subset of four representative tasks to isolate the contribution of each design choice. Results are summarized in Table~\ref{tab:ablation} and Table~\ref{tab:k_ablation}.

\begin{table}[t]
\centering
\caption{\textbf{Ablation studies (\textbf{Q2}).} Top: removing each auxiliary loss. Bottom: alternative design choices for the keyframe target and the query token. Best results per task are in \textbf{bold}.}
\label{tab:ablation}
\resizebox{0.98\linewidth}{!}{%
\begin{tabular}{l cccc c}
\toprule
\multirow{2}{*}{Variant} & Stack & Place & Lift & Click & \multirow{2}{*}{Avg} \\
 & Blocks & Burger & Pot & Alarm & \\
\midrule
$\pi_{0.5}$ (baseline) & 42 & 64 & 22 & 63 & 47.8 \\
\midrule
\method{} (full)          & \textbf{46} & \textbf{76} & \textbf{36} & \textbf{76} & \textbf{58.5} \\
w/o Stage Supervision            & 39 & 70 & 30 & 74 & 53.3 \\
w/o Keyframe Supervision         & 41 & 69 & 34 & 73 & 54.3 \\
\midrule
Keyframe Target: $t{+}H$       & 35 & 75 & 25 & 65 & 50.0 \\
Query: last VL-prefix token & 32 & 72 & 30 & 69 & 50.8 \\
\bottomrule
\end{tabular}%
}
\end{table}

\begin{table}[t]
\centering
\caption{\textbf{Sensitivity of stage-label expansion margin $K$ (\textbf{Q2}).} $K{=}25$ yields the best average.}
\label{tab:k_ablation}
\begin{tabular}{c cc c}
\toprule
$K$ & Place Burger & Adjust Bottle & Avg \\
\midrule
0  & 73 & 93 & 83.0 \\
15 & 68 & 94 & 81.0 \\
20 & 72 & 95 & 83.5 \\
\textbf{25} & \textbf{76} & \textbf{96} & \textbf{86.0} \\
30 & 73 & 94 & 83.5 \\
\bottomrule
\end{tabular}
\end{table}

\begin{figure*}[t]
  \centering
  \includegraphics[width=0.95\textwidth]{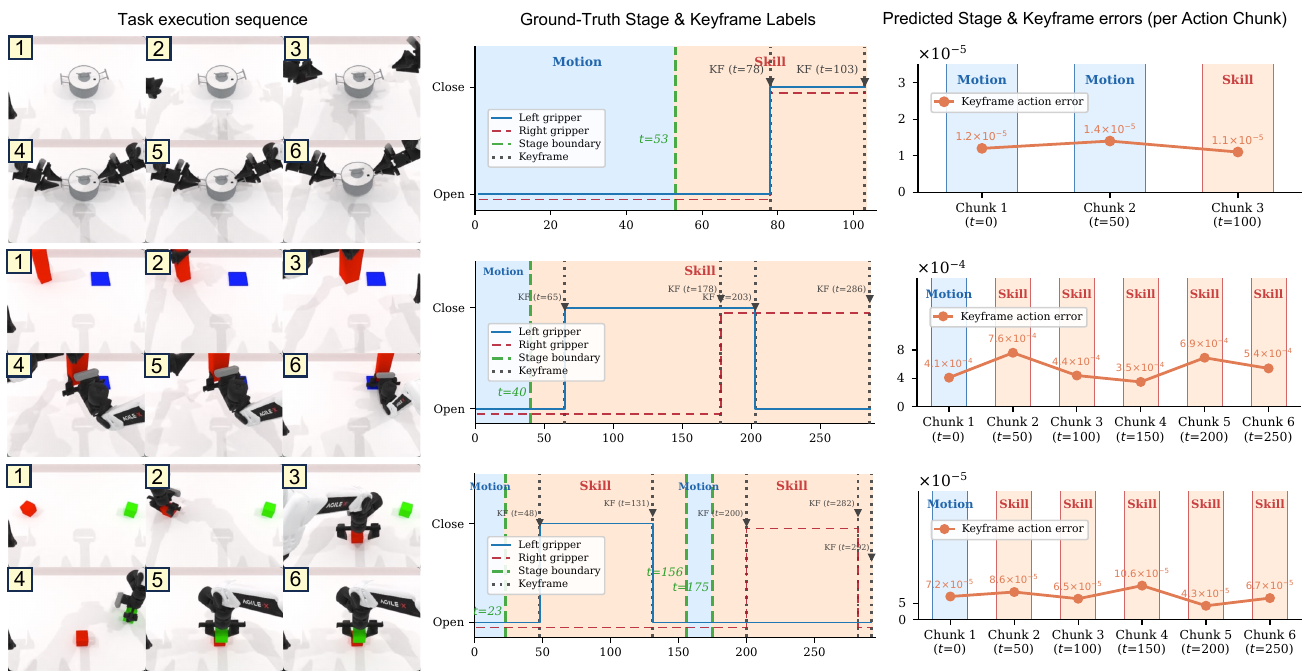}
  \caption{\textbf{Qualitative analysis (\textbf{Q3}).} Three tasks of increasing complexity (rows, top to bottom): \emph{Lift Pot} (2~KF), \emph{Handover Block} (4~KF), and \emph{Stack Blocks Two} (5~KF). Left column: task execution RGB sequence. Middle column: ground-truth dual-arm gripper states with annotated stage boundaries (green dashed) and keyframe positions (gray dotted). Right column: per-action-chunk predicted stage label (bar color: blue = Motion, orange = Skill) and keyframe prediction L1 error (orange line), reported in normalized joint space relative to each joint's feasible range ($<2\pi$).}
  \label{fig:mechanism}
\end{figure*}

\noindent
\textbf{Both auxiliary losses are necessary.}
Removing either component degrades performance: dropping stage supervision reduces the average from 58.5 to 54.3, and dropping keyframe supervision reduces it to 53.3. Neither alone recovers the full gain, confirming that the two losses provide complementary signals. Stage supervision captures the coarse motion/skill stage structure, while keyframe supervision anchors the policy to precise gripper-transition configurations.

\noindent
\textbf{Event-driven keyframe targets outperform fixed-horizon targets.}
Replacing the next-transition keyframe target with a fixed-horizon target at $t{+}H$ ($H{=}50$, the action chunk length) reduces the average to 50.0, only marginally above $\pi_{0.5}$. This validates the event-driven design: gripper-transition frames carry richer supervisory information than an arbitrary future timestep, because they correspond to physically meaningful state changes.

\noindent
\textbf{Dedicated learnable tokens matter.}
Replacing the learnable query token with the last token of the vision-language prefix reduces the average to 50.8. The dedicated token allows the model to develop a specialized representation for auxiliary prediction without interfering with the features used for action generation.

\noindent
\textbf{Expansion margin $K$.}
Table~\ref{tab:k_ablation} varies the stage-label expansion margin $K$ on two tasks. $K{=}0$ (no expansion) misses the pre-contact preparation and post-contact settling phases, while overly large $K$ (30) dilutes the skill label by incorporating irrelevant free-motion frames. $K{=}25$ provides the best balance, and performance remains stable across a reasonable range ($20$--$30$), indicating that the method is not sensitive to precise tuning of this hyperparameter.

\subsection{Qualitative Analysis (\textbf{Q3})}

Fig.~\ref{fig:mechanism} visualizes the auxiliary predictions during successful rollouts on three tasks of increasing complexity. Each row presents one task: the left column shows the RGB execution sequence, the middle column displays ground-truth gripper states with stage and keyframe annotations, and the right column reports per-action-chunk stage predictions and keyframe L1 errors. We highlight three observations.

\noindent
\textbf{Stage predictions faithfully track manipulation phases.}
Across all three tasks, the predicted stage label aligns with the ground-truth motion/skill boundary. In \emph{Lift Pot}, the model correctly predicts Motion for the first two chunks and switches to Skill at the third, matching the transition at $t{=}53$. In \emph{Handover Block} and \emph{Stack Blocks Two}, the model identifies Skill from the second chunk onward, consistent with the early stage boundary in both tasks. This indicates that the stage head has learned to anticipate upcoming contact phases rather than merely reacting to gripper closure.

\noindent
\textbf{Keyframe predictions remain accurate throughout the rollout.}
The keyframe L1 error stays consistently low across all action chunks: on the order of $10^{-5}$ for \emph{Lift Pot} and \emph{Stack Blocks Two}, and $10^{-4}$ for \emph{Handover Block}. Even on \emph{Stack Blocks Two}, where the keyframe target shifts five times, the error does not diverge. This confirms that the keyframe head maintains a reliable estimate of the next gripper-transition configuration regardless of task length.

\noindent
\textbf{Consistent behavior across task complexity.}
Both patterns hold from the 2-keyframe to the 5-keyframe task, suggesting that the auxiliary representations scale gracefully with the number of manipulation stages. This corroborates the finding in Table~\ref{tab:sim_results} that \method{} yields larger gains on multi-keyframe tasks, as the stage and keyframe heads provide an internal anchor at every gripper-transition window.

\begin{figure*}[t]
  \vspace{2pt}
  \centering
  \includegraphics[width=0.95\textwidth]{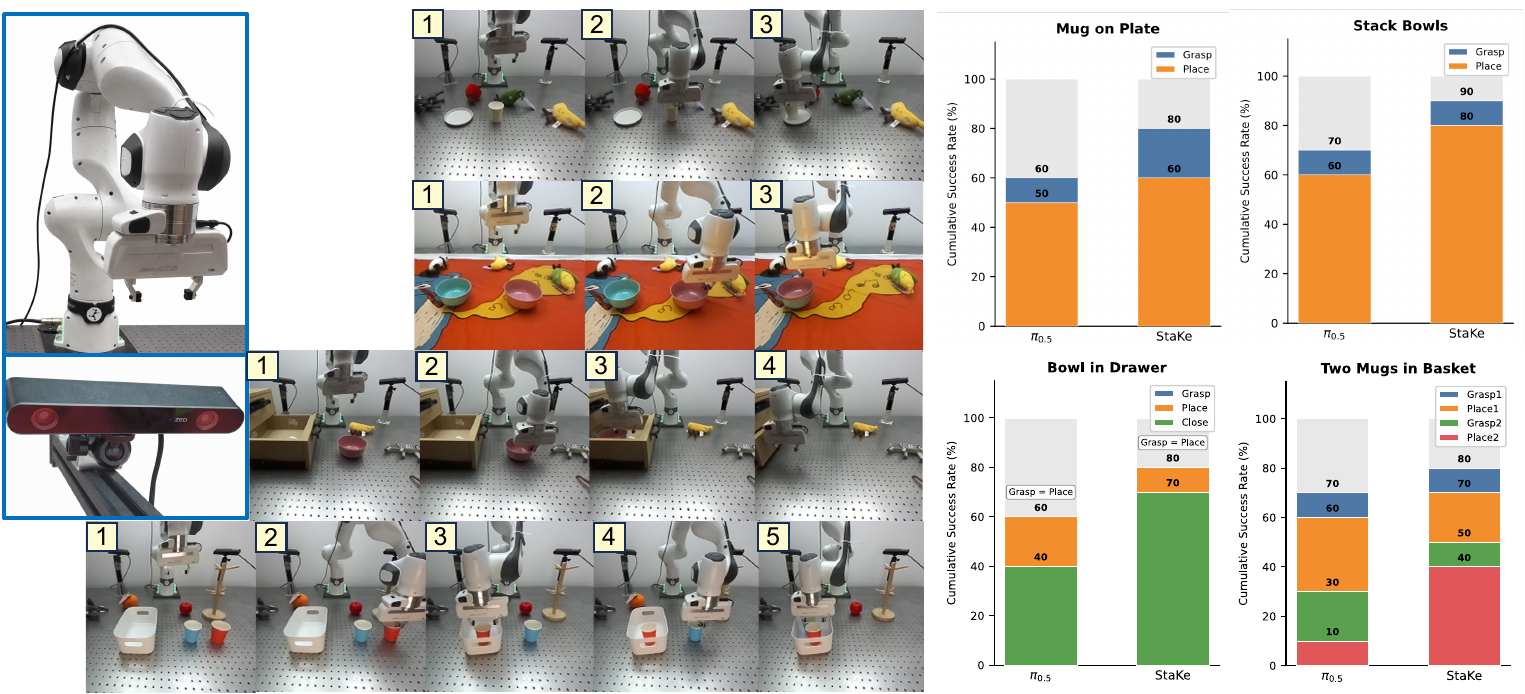}
  \caption{\textbf{Real-robot experiments (\textbf{Q4}).} Upper left: workspace with a Franka Research~3 arm and a front-view ZED~2i camera. Middle: start frame and keyframe snapshots for each task (top to bottom: Mug on Plate, Stack Bowls, Bowl in Drawer, Two Mugs in Basket). Right: cumulative success rate at each keyframe over 10 trials for $\pi_{0.5}$ and \method{}.}
  \label{fig:realrobot}
\end{figure*}

\noindent
\textbf{Consistent behavior across task complexity.}
Both patterns hold from the 2-keyframe to the 5-keyframe task, suggesting the auxiliary representations scale gracefully with the number of manipulation stages. This corroborates the finding in Table~\ref{tab:sim_results} that \method{} yields larger gains on multi-keyframe tasks, as the heads provide an internal anchor at every gripper-transition window.


\subsection{Real-Robot Experiments (\textbf{Q4})}


\noindent
\textbf{System setup.}
We evaluate on a Franka Research~3 arm with a parallel-jaw gripper, observed by a front-view ZED~2i RGB camera (Fig.~\ref{fig:realrobot}, upper left). Four tabletop tasks of increasing keyframe complexity are designed: \emph{put the white mug on the plate} (Mug on Plate, 2~KF: grasp, place), \emph{stack the red bowl on the green bowl} (Stack Bowls, 2~KF: grasp, place), \emph{put the red bowl in the drawer and close the drawer} (Bowl in Drawer, 3~KF: grasp, place, close), and \emph{put both the red mug and the blue mug in the basket} (Two Mugs in Basket, 4~KF: grasp, place, grasp, place). We collect 50 teleoperated demonstrations per task and train under a multi-task setting with the same hyperparameters as in Sec.~\ref{subsec:setup}. Each task is evaluated over 10 trials, and we report the cumulative success rate at each keyframe, where success at keyframe~$k$ requires completing all preceding keyframes.

\noindent
\textbf{Quantitative results.}
Fig.~\ref{fig:realrobot} (right) reports the cumulative success rates at each keyframe. \method{} improves the final task success on all four tasks, raising the average from 40.0\% ($\pi_{0.5}$) to 62.5\%. We highlight three findings.

\textbf{Consistent improvement across all tasks.}
\method{} outperforms $\pi_{0.5}$ on every task, with gains ranging from $+10\%$ on the simpler 2-keyframe tasks to $+30\%$ on the more demanding 3- and 4-keyframe tasks. This trend suggests that the auxiliary supervision becomes increasingly valuable as task complexity grows, because longer manipulation sequences expose the policy to more gripper-event transitions where keyframe-aware guidance is most needed.

\textbf{Gains concentrate at later keyframes.}
Both methods achieve comparable success at the initial grasp, but diverge at subsequent stages. For instance, in \emph{Bowl in Drawer}, $\pi_{0.5}$ drops from 60\% to 40\% at the final close action, while \method{} retains 70\%. This indicates that the stage head helps the policy recognize stage transitions such as placing-to-closing, where $\pi_{0.5}$ tends to hesitate.

\textbf{Robust to longer-horizon tasks.}
On \emph{Two Mugs in Basket} (4~KF), $\pi_{0.5}$ degrades sharply beyond the second grasp-place keyframe, suggesting insufficient awareness of the current task stage. The keyframe head of \method{} addresses this by providing an explicit next-transition target at every action chunk, sustaining execution across successive stages.

\section{CONCLUSION}

This paper introduced \method{}, a plug-in auxiliary supervision framework that improves VLA fine-tuning by injecting stage classification and keyframe prediction signals derived automatically from demonstrations, without modifying the inference loop. Experiments in both simulation and real-robot settings demonstrate consistent improvements across tasks of varying complexity, with ablation studies on auxiliary supervision and architectural design validating each component. Qualitative analysis further reveals that the learned representations faithfully track manipulation phases and provide accurate transition targets, explaining the source of performance gains. These results suggest that structured supervision is an effective and general strategy for enhancing VLA fine-tuning in long-horizon manipulation.

\noindent
\textbf{Limitations and Future Work.}
Stages and keyframes derive from binary gripper open/close events, so \method{} targets prehensile pick-and-place and does not directly extend to non-prehensile or dexterous manipulation. Future work will explore richer event definitions beyond binary gripper state and investigate leveraging the auxiliary representations at inference time.

\section{ACKNOWLEDGEMENT}

This work was jointly supported by National Natural Science Foundation of China (62322607, 62236010 and 62276261), Beijing Natural Science Foundation (L252033).








\bibliographystyle{IEEEtran}
\bibliography{root}

\begin{thebibliography}{10}
\providecommand{\url}[1]{#1}
\csname url@rmstyle\endcsname
\providecommand{\newblock}{\relax}
\providecommand{\bibinfo}[2]{#2}
\providecommand\BIBentrySTDinterwordspacing{\spaceskip=0pt\relax}
\providecommand\BIBentryALTinterwordstretchfactor{4}
\providecommand\BIBentryALTinterwordspacing{\spaceskip=\fontdimen2\font plus
\BIBentryALTinterwordstretchfactor\fontdimen3\font minus
  \fontdimen4\font\relax}
\providecommand\BIBforeignlanguage[2]{{%
\expandafter\ifx\csname l@#1\endcsname\relax
\typeout{** WARNING: IEEEtran.bst: No hyphenation pattern has been}%
\typeout{** loaded for the language `#1'. Using the pattern for}%
\typeout{** the default language instead.}%
\else
\language=\csname l@#1\endcsname
\fi
#2}}

\bibitem{zhu2024vision}
Y.~Ma, Z.~Song, Y.~Zhuang, J.~Hao, and I.~King, ``A survey on
  vision-language-action models for embodied ai,'' \emph{arXiv preprint
  arXiv:2405.14093}, 2024.

\bibitem{awadalla2023openflamingo}
A.~Awadalla, I.~Gao, J.~Gardner, J.~Hessel, Y.~Hanafy, W.~Zhu, K.~Marathe,
  Y.~Bitton, S.~Gadre, S.~Sagawa, \emph{et~al.}, ``Openflamingo: An open-source
  framework for training large autoregressive vision-language models,''
  \emph{arXiv preprint arXiv:2308.01390}, 2023.

\bibitem{liu2024visual}
H.~Liu, C.~Li, Q.~Wu, and Y.~J. Lee, ``Visual instruction tuning,''
  \emph{Advances in neural information processing systems}, vol.~36, pp.
  34\,892--34\,916, 2023.

\bibitem{brohan2022rt1}
A.~Brohan, N.~Brown, J.~Carbajal, Y.~Chebotar, J.~Dabis, C.~Finn,
  K.~Gopalakrishnan, K.~Hausman, A.~Herzog, J.~Hsu, \emph{et~al.}, ``Rt-1:
  Robotics transformer for real-world control at scale,'' \emph{arXiv preprint
  arXiv:2212.06817}, 2022.

\bibitem{brohan2023rt2}
A.~Brohan, N.~Brown, J.~Carbajal, Y.~Chebotar, X.~Chen, K.~Choromanski,
  T.~Ding, D.~Driess, A.~Dubey, C.~Finn, \emph{et~al.}, ``Rt-2:
  Vision-language-action models transfer web knowledge to robotic control,
  2023,'' \emph{URL https://arxiv. org/abs/2307.15818}, vol.~1, p.~2, 2024.

\bibitem{driess2023palme}
D.~Driess, F.~Xia, M.~S. Sajjadi, C.~Lynch, A.~Chowdhery, B.~Ichter, A.~Wahid,
  J.~Tompson, Q.~Vuong, T.~Yu, \emph{et~al.}, ``Palm-e: An embodied multimodal
  language model,'' \emph{arXiv preprint arXiv:2303.03378}, 2023.

\bibitem{kim2024openvla}
M.~J. Kim, K.~Pertsch, S.~Karamcheti, T.~Xiao, A.~Balakrishna, S.~Nair,
  R.~Rafailov, E.~Foster, G.~Lam, P.~Sanketi, \emph{et~al.}, ``Openvla: An
  open-source vision-language-action model, 2024,'' \emph{URL https://arxiv.
  org/abs/2406.09246}, vol.~1, no.~2, p.~4, 2024.

\bibitem{octo2024}
O.~M. Team, D.~Ghosh, H.~Walke, K.~Pertsch, K.~Black, O.~Mees, S.~Dasari,
  J.~Hejna, T.~Kreiman, C.~Xu, \emph{et~al.}, ``Octo: An open-source generalist
  robot policy,'' \emph{arXiv preprint arXiv:2405.12213}, 2024.

\bibitem{o2024open}
Q.~Vuong, S.~Levine, H.~R. Walke, K.~Pertsch, A.~Singh, R.~Doshi, C.~Xu,
  J.~Luo, L.~Tan, D.~Shah, \emph{et~al.}, ``Open x-embodiment: Robotic learning
  datasets and rt-x models,'' 2023.

\bibitem{fang2024rh20t}
H.-S. Fang, H.~Fang, Z.~Tang, J.~Liu, C.~Wang, J.~Wang, H.~Zhu, and C.~Lu,
  ``Rh20t: A comprehensive robotic dataset for learning diverse skills in
  one-shot,'' \emph{arXiv preprint arXiv:2307.00595}, 2023.

\bibitem{black2024pi0}
K.~Black, N.~Brown, D.~Driess, A.~Esmail, M.~Equi, C.~Finn, N.~Fusai, L.~Groom,
  K.~Hausman, B.~Ichter, \emph{et~al.}, ``{$\pi_{0}$: A Vision-Language-Action
  Flow Model for General Robot Control},'' \emph{arXiv preprint
  arXiv:2410.24164}, 2024.

\bibitem{physicalintelligence2025pi05}
P.~Intelligence, K.~Black, N.~Brown, J.~Darpinian, K.~Dhabalia, D.~Driess,
  A.~Esmail, M.~Equi, C.~Finn, N.~Fusai, \emph{et~al.}, ``{$\pi_{0.5}$: A
  Vision-Language-Action Model with Open-World Generalization},'' \emph{arXiv
  preprint arXiv:2504.16054}, 2025.

\bibitem{openvla_oft2025}
M.~J. Kim, C.~Finn, and P.~Liang, ``Fine-tuning vision-language-action models:
  Optimizing speed and success,'' \emph{arXiv preprint arXiv:2502.19645}, 2025.

\bibitem{bharadhwaj2023roboagent}
H.~Bharadhwaj, J.~Vakil, M.~Sharma, A.~Gupta, S.~Tulsiani, and V.~Kumar,
  ``Roboagent: Generalization and efficiency in robot manipulation via semantic
  augmentations and action chunking,'' pp. 4788--4795, 2024.

\bibitem{chen2025robotwin}
T.~Chen, Z.~Chen, B.~Chen, Z.~Cai, Y.~Liu, Z.~Li, Q.~Liang, X.~Lin, Y.~Ge,
  Z.~Gu, \emph{et~al.}, ``Robotwin 2.0: A scalable data generator and benchmark
  with strong domain randomization for robust bimanual robotic manipulation,''
  \emph{arXiv preprint arXiv:2506.18088}, 2025.

\bibitem{li2024roboflamingo}
X.~Li, M.~Liu, H.~Zhang, C.~Yu, J.~Xu, H.~Wu, C.~Cheang, Y.~Jing, W.~Zhang,
  H.~Liu, \emph{et~al.}, ``Vision-language foundation models as effective robot
  imitators,'' 2023.

\bibitem{li2024cogact}
Q.~Li, Y.~Liang, Z.~Wang, L.~Luo, X.~Chen, M.~Liao, F.~Wei, Y.~Deng, S.~Xu,
  Y.~Zhang, \emph{et~al.}, ``Cogact: A foundational vision-language-action
  model for synergizing cognition and action in robotic manipulation,''
  \emph{arXiv preprint arXiv:2411.19650}, 2024.

\bibitem{liu2024rdt}
S.~Liu, L.~Wu, B.~Li, H.~Tan, H.~Chen, Z.~Wang, K.~Xu, H.~Su, and J.~Zhu,
  ``Rdt-1b: a diffusion foundation model for bimanual manipulation,''
  \emph{arXiv preprint arXiv:2410.07864}, 2024.

\bibitem{cheang2024gr2}
C.-L. Cheang, G.~Chen, Y.~Jing, T.~Kong, H.~Li, Y.~Li, Y.~Liu, H.~Wu, J.~Xu,
  Y.~Yang, \emph{et~al.}, ``Gr-2: A generative video-language-action model with
  web-scale knowledge for robot manipulation,'' \emph{arXiv preprint
  arXiv:2410.06158}, 2024.

\bibitem{wang2024hpt}
L.~Wang, X.~Chen, J.~Zhao, and K.~He, ``Scaling proprioceptive-visual learning
  with heterogeneous pre-trained transformers,'' vol.~37, 2024, pp.
  124\,420--124\,450.

\bibitem{qu2025spatialvla}
D.~Qu, H.~Song, Q.~Chen, Y.~Yao, X.~Ye, Y.~Ding, Z.~Wang, J.~Gu, B.~Zhao,
  D.~Wang, \emph{et~al.}, ``Spatialvla: Exploring spatial representations for
  visual-language-action model,'' \emph{arXiv preprint arXiv:2501.15830}, 2025.

\bibitem{liu2025hybridvla}
J.~Liu, H.~Chen, P.~An, Z.~Liu, R.~Zhang, C.~Gu, X.~Li, Z.~Guo, S.~Chen,
  M.~Liu, \emph{et~al.}, ``Hybridvla: Collaborative diffusion and
  autoregression in a unified vision-language-action model,'' \emph{arXiv
  preprint arXiv:2503.10631}, 2025.

\bibitem{pertsch2025fast}
K.~Pertsch, K.~Stachowicz, B.~Ichter, D.~Driess, S.~Nair, Q.~Vuong, O.~Mees,
  C.~Finn, and S.~Levine, ``Fast: Efficient action tokenization for
  vision-language-action models,'' \emph{arXiv preprint arXiv:2501.09747},
  2025.

\bibitem{wen2024tinyvla}
J.~Wen, Y.~Zhu, J.~Li, M.~Zhu, Z.~Tang, K.~Wu, Z.~Xu, N.~Liu, R.~Cheng,
  C.~Shen, \emph{et~al.}, ``Tinyvla: Towards fast, data-efficient
  vision-language-action models for robotic manipulation,'' \emph{IEEE Robotics
  and Automation Letters}, 2025.

\bibitem{yue2024deervla}
Y.~Yue, Y.~Wang, B.~Kang, Y.~Han, S.~Wang, S.~Song, J.~Feng, and G.~Huang,
  ``Deer-vla: Dynamic inference of multimodal large language models for
  efficient robot execution,'' vol.~37, 2024, pp. 56\,619--56\,643.

\bibitem{chen2025conrft}
Y.~Chen, S.~Tian, S.~Liu, Y.~Zhou, H.~Li, and D.~Zhao, ``Conrft: A reinforced
  fine-tuning method for vla models via consistency policy,'' \emph{arXiv
  preprint arXiv:2502.05450}, 2025.

\bibitem{guo2025irevla}
Y.~Guo, J.~Zhang, X.~Chen, X.~Ji, Y.-J. Wang, Y.~Hu, and J.~Chen, ``Improving
  vision-language-action model with online reinforcement learning,'' pp.
  15\,665--15\,672, 2025.

\bibitem{shridhar2023perceiver}
M.~Shridhar, L.~Manuelli, and D.~Fox, ``Perceiver-actor: A multi-task
  transformer for robotic manipulation,'' pp. 785--799, 2023.

\bibitem{goyal2023rvt}
A.~Goyal, J.~Xu, Y.~Guo, V.~Blukis, Y.-W. Chao, and D.~Fox, ``Rvt: Robotic view
  transformer for 3d object manipulation,'' pp. 694--710, 2023.

\bibitem{gervet2023act3d}
T.~Gervet, Z.~Xian, N.~Gkanatsios, and K.~Fragkiadaki, ``Act3d: 3d feature
  field transformers for multi-task robotic manipulation,'' 2023.

\bibitem{chen2025c2fkeyframe}
X.~Zhu, D.~Klee, D.~Wang, B.~Hu, H.~Huang, A.~Tangri, R.~Walters, and R.~Platt,
  ``Coarse-to-fine 3d keyframe transporter,'' \emph{arXiv preprint
  arXiv:2502.01773}, 2025.

\bibitem{shi2023awe}
L.~X. Shi, A.~Sharma, T.~Z. Zhao, and C.~Finn, ``Waypoint-based imitation
  learning for robotic manipulation,'' \emph{arXiv preprint arXiv:2307.14326},
  2023.

\bibitem{zhang2024pivotr}
K.~Zhang, P.~Ren, B.~Lin, J.~Lin, S.~Ma, H.~Xu, and X.~Liang, ``Pivot-r:
  Primitive-driven waypoint-aware world model for robotic manipulation,''
  vol.~37, 2024, pp. 54\,105--54\,136.

\bibitem{dipalo2024kat}
N.~Di~Palo and E.~Johns, ``Keypoint action tokens enable in-context imitation
  learning in robotics,'' \emph{arXiv preprint arXiv:2403.19578}, 2024.

\bibitem{huang2024rekep}
W.~Huang, C.~Wang, Y.~Li, R.~Zhang, and L.~R. Fei-Fei, ``Spatio-temporal
  reasoning of relational keypoint constraints for robotic manipulation,''
  \emph{arXiv preprint arXiv:2409.01652}, vol.~2, 2024.

\bibitem{kou2024kisa}
L.~Kou, F.~Ni, Y.~Zheng, J.~Liu, Y.~Yuan, Z.~Dong, and J.~Hao, ``Kisa: A
  unified keyframe identifier and skill annotator for long-horizon robotics
  demonstrations,'' in \emph{Forty-first International Conference on Machine
  Learning}, 2024.

\bibitem{prime2024}
T.~Gao, S.~Nasiriany, H.~Liu, Q.~Yang, and Y.~Zhu, ``Prime: Scaffolding
  manipulation tasks with behavior primitives for data-efficient imitation
  learning,'' \emph{IEEE Robotics and Automation Letters}, vol.~9, no.~10, pp.
  8322--8329, 2024.

\bibitem{atomskill2025}
Y.~Zhu, W.~Wang, S.~Wu, Y.~Shi, and J.~Wang, ``Learning semantic atomic skills
  for multi-task robotic manipulation,'' \emph{arXiv preprint
  arXiv:2512.18368}, 2025.

\bibitem{dexskills2024}
X.~Mao, G.~Giudici, C.~Coppola, K.~Althoefer, I.~Farkhatdinov, Z.~Li, and
  L.~Jamone, ``Dexskills: Skill segmentation using haptic data for learning
  autonomous long-horizon robotic manipulation tasks,'' in \emph{2024 IEEE/RSJ
  international conference on intelligent robots and systems (IROS)}.\hskip 1em
  plus 0.5em minus 0.4em\relax IEEE, 2024, pp. 5104--5111.

\bibitem{ma2024sigmaagent}
T.~Ma, J.~Zhou, Z.~Wang, R.~Qiu, and J.~Liang, ``Contrastive imitation learning
  for language-guided multi-task robotic manipulation,'' \emph{arXiv preprint
  arXiv:2406.09738}, 2024.

\bibitem{chi2023diffusion}
C.~Chi, Z.~Xu, S.~Feng, E.~Cousineau, Y.~Du, B.~Burchfiel, R.~Tedrake, and
  S.~Song, ``Diffusion policy: Visuomotor policy learning via action
  diffusion,'' \emph{The International Journal of Robotics Research}, vol.~44,
  no. 10-11, pp. 1684--1704, 2025.

\bibitem{zhao2023act}
T.~Z. Zhao, V.~Kumar, S.~Levine, and C.~Finn, ``Learning fine-grained bimanual
  manipulation with low-cost hardware,'' \emph{arXiv preprint
  arXiv:2304.13705}, 2023.

\end{thebibliography}



\end{document}